\documentclass[pdflatex,sn-basic,Numbered]{sn-jnl} 

\usepackage{graphicx}
\usepackage{amsmath,amssymb,amsfonts}
\usepackage{booktabs}
\usepackage{xcolor}
\usepackage{manyfoot} 

\makeatletter
\gdef\orcidlogo{\textcolor[HTML]{A6CE39}{\scriptsize\textbf{iD}}}
\makeatother

\begin{document}

\title[Unsupervised Event-Based Pedestrian Crossing Detection]{Unsupervised spiking feature learning for event-based pedestrian crossing detection: approaching supervised accuracy without labelled training data}

\author*[1]{\fnm{Henok} \sur{Teklu}\,\orcid{https://orcid.org/0009-0003-7890-7200}}\email{henok.teklu@almamater.si}
\author[2]{\fnm{Mustafa} \sur{Sakhai}\,\orcid{https://orcid.org/0000-0003-3941-6065}}\email{msakhai@agh.edu.pl}
\author[1]{\fnm{Matej} \sur{Mertik}\,\orcid{https://orcid.org/0000-0002-4557-330X}}\email{matej.mertik@almamater.si}
\author*[2]{\fnm{Maciej} \sur{Wielgosz}\,\orcid{https://orcid.org/0000-0002-4401-2957}}\email{wielgosz@agh.edu.pl}
\affil*[1]{\orgdiv{Department of Applied Artificial Intelligence}, \orgname{Alma Mater Europaea University}, \orgaddress{\city{Maribor}, \country{Slovenia}}}
\affil[2]{\orgname{AGH University of Science and Technology}, \orgaddress{\city{Krak\'ow}, \country{Poland}}}

\abstract{Event cameras are well suited to pedestrian crossing
detection, and spiking neural networks (SNNs) can process their
output natively, but current SNN detectors are trained with
supervised backpropagation and therefore require costly frame-level
crossing labels. We
investigate crossing detection with no labels in feature learning
and report the first unsupervised results on the recent
DVS-PedX pedestrian-crossing benchmark. A single spiking layer
trained with winner-take-all spike-timing-dependent plasticity learns
a dictionary from unlabelled event-frame patches; frames are encoded
by cosine similarity to the learned filters with spatial pooling and
read out by a linear classifier, the only supervised
component. On the 24{,}454-frame test
split the method attains 90.3\% accuracy and an area under the
receiver operating characteristic curve (AUROC) of 0.936, compared
with 92.0\% and 0.943 for a supervised spiking network trained
end-to-end on the same frames; under adverse weather the AUROC is
0.913. The result is insensitive to the choice of plasticity rule
but depends strongly on the readout protocol: with the classical
neuron-assignment readout the same network attains only 0.70 AUROC.
On the benchmark's real converted portion, a readout refit lifts
performance from chance (0.53) to 0.67 AUROC, within the published
supervised range. These
findings indicate that the accuracy cost of removing labels from
feature learning is small on this benchmark, and that reported
weaknesses of unsupervised spiking networks may be attributable to
the readout protocol rather than to the learning rule.}


\maketitle

\section{Introduction}\label{sec:intro}

Detecting whether a pedestrian is about to cross the road is a
safety-critical task for driving perception systems
\cite{rasouli2020,dollar2012}. Event cameras
\cite{lichtsteiner2008,posch2011,brandli2014} offer microsecond
latency, high dynamic range and freedom from motion blur
\cite{gallego2022}, properties that make them attractive sensors for
this task. Spiking neural networks (SNNs)
\cite{maass1997,gerstner2002} process event streams natively and can
be executed efficiently on neuromorphic hardware
\cite{davies2018,davies2021,merolla2014,furber2014,pei2019}.
Supervised SNNs trained with surrogate-gradient methods
\cite{neftci2019,wu2018stbp,fang2023spikingjelly} have achieved high
crossing-detection accuracy on simulated and converted event data,
including under adverse weather
\cite{sakhai2024,sakhai2026dvspedx}. Such training,
however, requires frame-level crossing annotations, which are slow
and expensive to produce and are often behaviourally ambiguous.
The conditions under which detection is most difficult, such as
night and heavy rain, are also those for which labelled examples are
scarcest.

In this paper we examine how accurately crossing detection can be
performed when the feature-learning stage uses no labels at all. On
the DVS-PedX benchmark \cite{sakhai2026dvspedx} we find that the
resulting accuracy falls within a fraction of a point of that
obtained with full supervision. The contributions of this paper are
as follows.
\begin{enumerate}
\item We report the first unsupervised results on DVS-PedX: 90.3\%
accuracy and 0.936 area under the receiver operating characteristic
curve (AUROC), compared with 92.0\% and 0.943 for a supervised SNN
trained end-to-end on the same frames, and an adverse-weather AUROC
of 0.913 (Sec.~\ref{sec:results-headline}). Labels are used only to
fit a linear readout and for evaluation.
\item We present a readout analysis showing that the same learned
network spans an AUROC range of 0.70 to 0.94 depending only on how
its output code is read out (Sec.~\ref{sec:results-tiers}), which
indicates that the classical neuron-assignment protocol, rather than
the local learning rule, limits reported performance.
\item We provide robustness analyses: the result is insensitive to
the plasticity rule (a novelty-modulated dictionary differs by less
than 0.001 AUROC), and we report per-weather performance and an
operating-characteristic analysis for each weather partition
(Sec.~\ref{sec:results-operating}).
\item We report label-free results on the benchmark's real converted
portion under three training regimes (zero-shot transfer from
simulation, readout adaptation, and training on the real data
itself), with evaluation partitioned by the weather attribute of
each scene (Sec.~\ref{sec:results-jaad}).
\end{enumerate}

The remainder of the paper is organised as follows.
Section~\ref{sec:related} positions the work in the literature;
Sec.~\ref{sec:method} describes the pipeline;
Sec.~\ref{sec:setup} the data, references and metrics;
Sec.~\ref{sec:results} the results;
Sec.~\ref{sec:discussion} discusses implications, limitations and
future work; and Sec.~\ref{sec:conclusion} concludes. Appendices
list all hyperparameters and protocol details for reproduction.

\section{Related work}\label{sec:related}

\paragraph{Crossing intent from conventional video.} Pedestrian
detection was developed largely on frame-based benchmarks
\cite{dollar2012} and on driving-scene datasets such as KITTI and
Cityscapes \cite{geiger2013,cordts2016}. The JAAD \cite{rasouli2017}
and PIE \cite{rasouli2019pie} datasets later established the recognition
of crossing \emph{behaviour} as a benchmark problem in its own right;
see \cite{rasouli2020} for a survey. Deep video models perform well
in this setting but require dense annotation.

\paragraph{Event-based pedestrian perception.} Supervised SNNs have
been used to detect street-crossing from dynamic vision sensor
streams in simulation, with adverse weather serving as the stress
condition \cite{sakhai2024}, and event sensors have been
compared with RGB cameras with respect to robustness in driving
simulators \cite{sakhai2025sensors}. DVS-PedX \cite{sakhai2026dvspedx}
consolidated this line of work into what is, to our knowledge, the
first public benchmark dedicated to event-based crossing detection;
it pairs synthetic CARLA-simulated scenes under controlled weather
with real JAAD dash-cam clips \cite{rasouli2017} converted to events
using the v2e video-to-events tool \cite{hu2021} (JAAD-DVS), and
provides frame-level binary
labels. Its baseline experiments use supervised SNNs; in separate
work we applied a supervised convolutional SNN to the JAAD-DVS
crossing task and found that temporal data augmentation improved
classification accuracy \cite{teklu2026crossing}.
That network is a different model from the single-frame supervised
reference we train for comparison here (Sec.~\ref{sec:setup}), and no
label-free results had been reported on the benchmark prior to the
present work.

\paragraph{Unsupervised spiking feature learning.}
Spike-timing-dependent plasticity (STDP)
\cite{markram1997,bi1998,song2000,caporale2008} supports label-free
feature extraction from spike trains \cite{masquelier2007};
winner-take-all (WTA) layers with homeostatic thresholds
\cite{diehl2015,turrigiano2004} made it practical, and convolutional
hierarchies \cite{kheradpisheh2018}, reward-modulated and
three-factor variants \cite{mozafari2019,mazurek2025threefactor} and
soft-WTA Hebbian rules \cite{moraitis2022}
extend the family, with \cite{falez2019} quantifying the gap to
conventional feature learning \cite{olshausen1996,coates2011}. On event-camera
benchmarks, including digit and gesture recognition
\cite{orchard2015,amir2017,li2017cifar}, automotive recognition and
detection \cite{sironi2018,detournemire2020,perot2020}, and steering
prediction \cite{maqueda2018}, learned label-free features have
generally lagged behind handcrafted descriptors
\cite{lagorce2017,sironi2018,ramesh2020}
and sparse-coding dictionaries \cite{aharon2006,kostadinov2021}. In
the present work a WTA-STDP layer is used as a patch-dictionary
learner within a classical single-layer recognition pipeline
\cite{coates2011}, the approach is applied to a safety-relevant
task, and its performance is quantified against a supervised
reference on identical test data.

\section{Method}\label{sec:method}

The pipeline consists of four stages; labels are used only in the
last.

\paragraph{Representation.} DVS-PedX provides 33\,ms event-frame
accumulations. Each frame is a single-channel $64{\times}64$ count
image normalised to $[0,1]$ by its 99th percentile.

\begin{figure}[t]
\centering
\includegraphics[width=\textwidth]{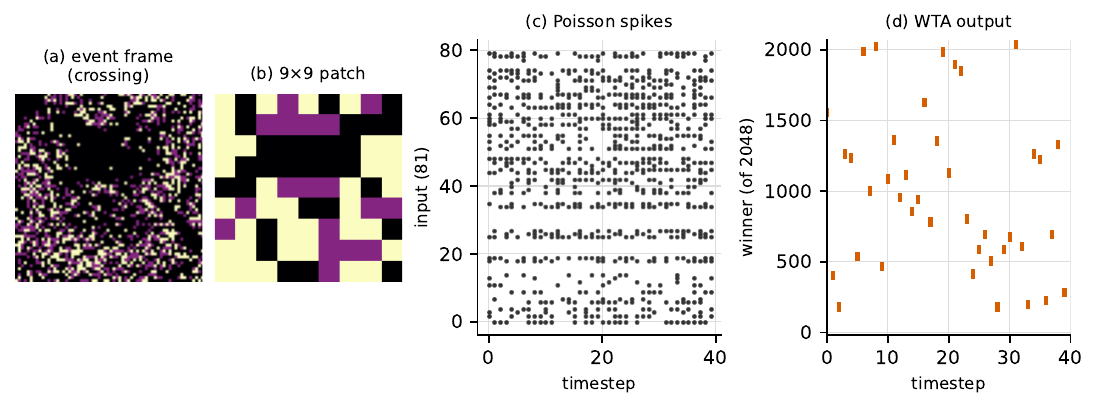}
\caption{Input path on one real test frame (crossing). (a) The
$64{\times}64$ event frame. (b) One $9{\times}9$ patch. (c) Poisson
spike trains generated from the patch (81 input lines, $T{=}40$
timesteps). (d) WTA response of the trained dictionary to
those spikes: a sparse sequence of winner identities, the code from
which both STDP updates (during learning) and encodings (at readout)
derive.}\label{fig:encoding}
\end{figure}

\paragraph{Unsupervised dictionary.} The learner is a single layer of
$K{=}2048$ leaky integrate-and-fire neurons \cite{gerstner2002} with non-negative weights
$W \in \mathbb{R}^{K \times 81}_{\ge 0}$, following \cite{diehl2015}.
A patch with normalised intensities $u \in [0,1]^{81}$ is presented for
$T{=}40$ timesteps as Poisson spike trains, $s_j(t) \sim
\mathrm{Bernoulli}(\lambda u_j)$. Membrane potentials integrate with
leak,
\begin{equation}
V_k(t) \;=\; e^{-1/\tau_v}\, V_k(t-1) \;+\; W_k\, s(t),
\qquad \tau_v = 20,
\end{equation}
and at each timestep the largest supra-threshold potential wins
(hard winner-take-all), spikes, resets, and subtractively inhibits the
rest; a per-neuron adaptive threshold offset $\theta_k$ grows with each
win and decays slowly, holding firing rates homeostatic. The winner
updates by weight-dependent trace STDP,
\begin{equation}
\Delta W_{k} \;=\; \eta_{+}\, \bar{x}\,(w_{\max}-W_k)\;-\;
\eta_{-}\,(1-\bar{x})\,W_k ,
\end{equation}
with $\bar{x}$ the presynaptic trace, followed by $\ell_1$
row-normalisation. None of the quantities above depends on class
labels. Training draws 300k random patches from the training frames
and takes eight minutes on a single consumer GPU (RTX 5060 Ti); the
complete hyperparameter set is listed in Appendix~\ref{app:hyper}.
Figure~\ref{fig:filters} shows the learned filters, which comprise
oriented edges, blobs and silhouette fragments;
Fig.~\ref{fig:encoding} illustrates the input path for one test
frame, from event frame to WTA output spikes.

\begin{figure}[t]
\centering
\includegraphics[width=0.8\textwidth]{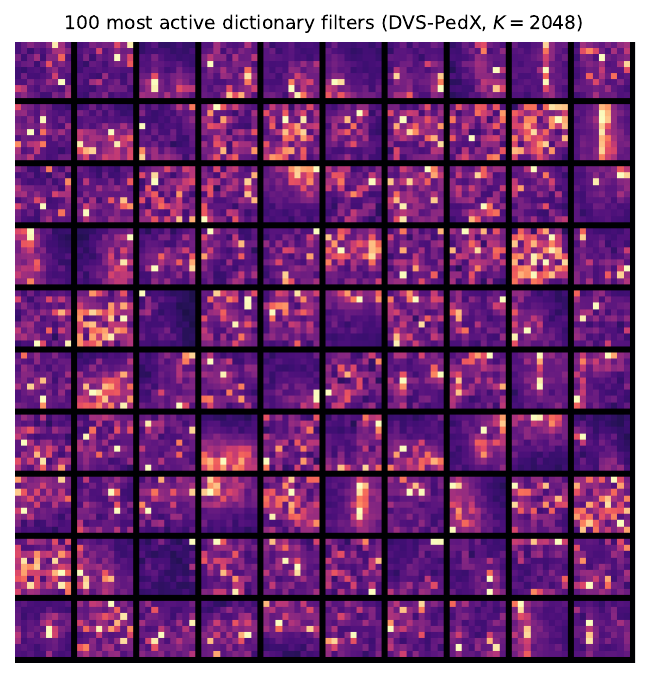}
\caption{Dictionary learned without labels from DVS-PedX event-frame
patches: oriented edges, blobs and silhouette fragments.}\label{fig:filters}
\end{figure}

\paragraph{Encoding and pooling.} Every $9{\times}9$ patch on a
stride-3 grid is encoded by cosine similarity to all filters with the
triangle activation $f_k = \max(0, r_k - \bar{r})$ \cite{coates2011},
and responses are sum-pooled over a $4{\times}4$ spatial grid,
yielding $16K$ features per frame.

\paragraph{Readout.} The readout is an $\ell_2$-regularised logistic
regression fitted on the frozen features. This is the only supervised
component, and it follows the protocol used by published event-vision
baselines for both handcrafted and learned features. For the readout
analysis of Sec.~\ref{sec:results-tiers} we additionally evaluate the
classical neuron--class assignment readout \cite{diehl2015} on a
whole-frame variant of the same learner.

\section{Experimental setup}\label{sec:setup}

\paragraph{Data.} DVS-PedX \cite{sakhai2026dvspedx} synthetic
portion: 163{,}012 single frames from CARLA sequences under normal
and adverse weather, binary crossing labels, stratified 70/15/15
split (test $n{=}24{,}454$), class prior $\approx$80\%
non-crossing. This is the same split and frame protocol as used for
the supervised reference, so that comparisons are made on identical
test data. The benchmark's real portion (JAAD-DVS) consists of the
346 dash-cam videos of the JAAD dataset \cite{rasouli2017} converted
to events with v2e \cite{hu2021}; 41{,}976 of its frames
carry binary crossing labels (38\% crossing), which we split into
34{,}567 training and 7{,}409 test frames under the same seed-42
protocol. Each JAAD video carries a weather attribute (clear,
cloudy, rain or snow) in its annotations, which we use to partition
the real test set by environment (Sec.~\ref{sec:results-jaad}).

\paragraph{Reference methods.} The majority-class baseline scores
80.6\% accuracy. As a matched supervised reference we trained a
single-frame SNN with adaptive modulation end-to-end with
backpropagation on
the same training split; it attains 92.0\% accuracy and 0.943 AUROC.
A same-split reference is necessary because published supervised
results on this benchmark's data use different protocols: clip-level
experiments on the precursor of the synthetic portion
\cite{sakhai2024} report, for event data under adverse weather,
0.950 AUROC for a spiking model (SPS R18T, nine-frame clips) and
0.988 AUROC for a non-spiking ResNet18, with all evaluated models
exceeding 0.94 AUROC under normal weather. These clip-level figures
on pre-release data provide published context but are not directly
comparable to the single-frame protocol used here.

\paragraph{Metrics.} Because the class prior is approximately 80/20,
accuracy alone can be misleading; we therefore report AUROC, balanced
accuracy, macro-F1 and per-class F1 throughout, together with
per-weather partitions.

\section{Results}\label{sec:results}

\subsection{Overall performance}\label{sec:results-headline}

\begin{figure}[t]
\centering
\includegraphics[width=\textwidth]{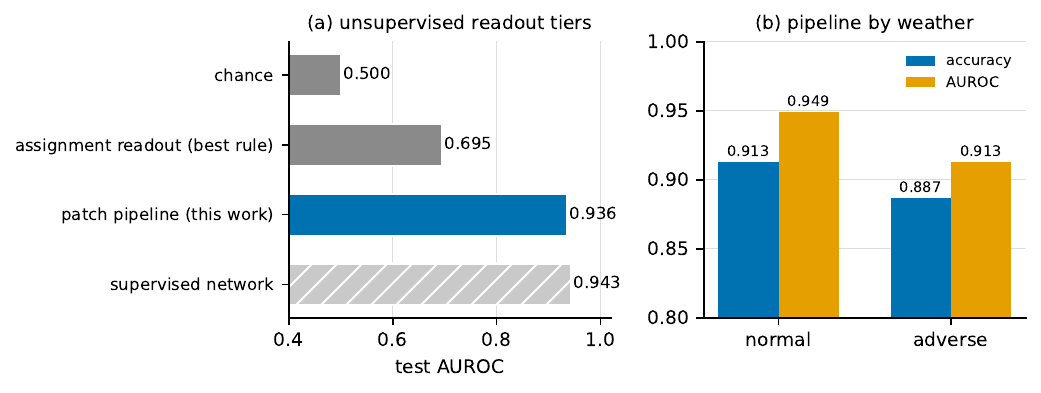}
\caption{(a) AUROC of the same unsupervised learner under two readout
protocols: neuron assignment (0.70) and the patch pipeline with a
linear probe (0.936), with features learned by the same local rule
in both cases. (b) Pipeline performance by weather
partition.}\label{fig:tiers}
\end{figure}

Table~\ref{tab:headline} summarises the main result. The unsupervised
pipeline reaches 90.3\% accuracy and 0.936 AUROC on the
24{,}454-frame test split, corresponding to 96\% of the accuracy
interval between the majority baseline and the supervised network,
and within 0.007 AUROC of the latter. The per-weather partitions
(Fig.~\ref{fig:tiers}b) give 91.3\% accuracy and 0.949 AUROC under
normal weather and 88.7\% and 0.913 under adverse weather. The
normal-weather AUROC exceeds the supervised model's combined value,
and the reduction under adverse weather (0.036 AUROC) is moderate.
This observation is relevant in practice because adverse conditions
are those for which labelled data is most difficult to obtain and
label-free training is therefore most useful. For published context,
supervised clip-level results on the precursor dataset lie in the
range 0.94--0.99 AUROC \cite{sakhai2024} (Sec.~\ref{sec:setup});
the unsupervised results reported here fall within the lower part of
this range despite using no labels for feature learning and a
single-frame protocol that provides less temporal context.

\begin{table}[h]
\caption{DVS-PedX synthetic test split ($n{=}24{,}454$). Labels are
used by the pipeline only in the linear readout.}\label{tab:headline}
\begin{tabular}{@{}lccccc@{}}
\toprule
Method & Feature labels & Acc.\ & AUROC & macro-F1 & F1 (crossing) \\
\midrule
Majority class & --- & 0.806 & 0.500 & 0.444 & 0.000 \\
Assignment readout (best rule) & none & 0.652 & 0.695 & 0.514 & 0.425 \\
\textbf{Patch pipeline (this work)} & none & \textbf{0.903} & \textbf{0.936} & \textbf{0.834} & \textbf{0.726} \\
\quad novelty-modulated variant & none & 0.903 & 0.937 & 0.834 & 0.727 \\
Supervised SNN (same split) & all & 0.920 & 0.943 & --- & --- \\
\bottomrule
\end{tabular}
\end{table}

\subsection{Effect of the readout protocol}\label{sec:results-tiers}

Unsupervised spiking networks have generally underperformed on event
benchmarks, and Fig.~\ref{fig:tiers}a indicates a likely reason for
this task. The same learning rule yields 0.695 AUROC under
whole-frame learning with the neuron-assignment readout, which is the
protocol of the classical literature, and 0.936 under the patch
dictionary with pooled encoding and a linear probe: a difference of
0.24 AUROC with no change to the plasticity mechanism. Conversely,
changing the plasticity rule has almost no effect on the pipeline: a
novelty-modulated three-factor variant
\cite{mazurek2025threefactor} of the dictionary differs from
plain STDP by less than 0.001 AUROC (Table~\ref{tab:headline}).
These observations indicate that the local learning rule is not the
limiting factor for this task, and that representation structure and
readout account for most of the performance difference.

Two consequences follow. First, for neuromorphic deployment,
hardware whose on-chip plasticity implements only pair-based STDP
loses little on this task relative to richer neuromodulated rules,
because the remaining stages (encoding, pooling and readout) are
computationally inexpensive and standard. Second, comparisons
between spiking learning rules that are mediated by weak readouts
may both under-report the rules and distort their relative
differences; in our experiments the rule comparison became
interpretable only once the readout was no longer the limiting
stage.

We also note that the two regimes differ with respect to the effect
of the rule. In the whole-frame regime the novelty-modulated rule
outperformed plain STDP consistently (three seeds of three, $+0.031$
AUROC), whereas in the patch regime the two are indistinguishable. A
plausible explanation lies in the scale of the learning unit: whole
frames carry class structure that renders some inputs rare and
novelty-gating informative, whereas local patches are largely
class-agnostic, so the gating signal has no systematic structure to
amplify. The relative merit of a learning rule therefore appears to
depend on the statistics of the unit to which it is applied.

\subsection{Effect of representation and dictionary capacity}\label{sec:results-ladder}

Table~\ref{tab:ladder} shows how performance depends on the
representation and on dictionary capacity under an identical readout
protocol. The change from whole-frame codes to a patch dictionary
accounts for the largest improvement (+0.18 AUROC, obtained with a
dictionary two orders of magnitude smaller), increasing the
dictionary size from $K{=}49$ to $K{=}2048$ adds a further 0.05
AUROC, and the choice of learning rule has no measurable effect. The
$K{=}49$ configuration is of independent interest: a dictionary of
only 49 filters, small enough to be inspected individually, reaches
87.6\% accuracy, approximately ten points above the best whole-frame
protocol.

\begin{table}[h]
\caption{Configuration ladder on the synthetic test split (identical
linear-probe readout except the first row).}\label{tab:ladder}
\begin{tabular}{@{}lccc@{}}
\toprule
Configuration & $K$ & Acc.\ & AUROC \\
\midrule
whole-frame, assignment readout (best rule) & 1600 & 0.652 & 0.695 \\
patch dictionary & 49 & 0.876 & 0.886 \\
patch dictionary & 2048 & \textbf{0.903} & \textbf{0.936} \\
\quad novelty-modulated variant & 2048 & 0.903 & 0.937 \\
\bottomrule
\end{tabular}
\end{table}

\subsection{Operating-point analysis}\label{sec:results-operating}

\begin{figure}[t]
\centering
\includegraphics[width=\textwidth]{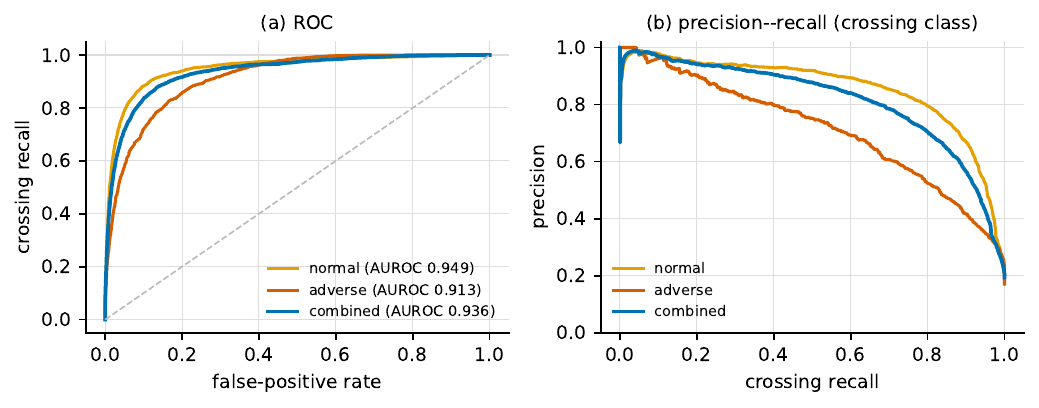}
\caption{Operating characteristics of the unsupervised pipeline on the
DVS-PedX synthetic test split, overall and per weather partition.
(a) ROC. (b) Precision--recall for the crossing class.}\label{fig:operating}
\end{figure}

Aggregate metrics summarise the score distribution, whereas a
deployed detector operates at a specific point on it; for a safety
function the relevant quantity is the crossing recall attainable
under a fixed false-alarm budget. Figure~\ref{fig:operating} shows
the ROC and precision--recall curves of the unsupervised pipeline on
the synthetic test split for each weather partition, and
Table~\ref{tab:operating} lists the corresponding operating points.
At a false-positive rate of 5\% the detector retains 71.6\% crossing
recall, and at 10\% it retains 83.5\%. Requiring 90\% precision on
crossing alerts reduces recall to 42.2\%, which indicates the
distance that remains between this detector and a deployable safety
function. The per-weather curves quantify the operational meaning of
the AUROC difference in Fig.~\ref{fig:tiers}b: adverse conditions
reduce recall at any fixed false-alarm budget, which motivates
condition-dependent thresholds and indicates where additional
(unlabelled) adverse training data would be most useful. Because the
score is continuous and the readout is linear, the operating point
can be recalibrated freely, including on a per-condition basis.

\begin{table}[h]
\caption{Operating points on the synthetic test split (combined
weather): crossing recall under fixed precision and
false-positive-rate constraints, computed from the readout
score.}\label{tab:operating}
\begin{tabular}{@{}lc@{}}
\toprule
Constraint & Crossing recall \\
\midrule
precision $\ge 0.8$ & 0.674 \\
precision $\ge 0.9$ & 0.422 \\
false-positive rate $\le 0.05$ & 0.716 \\
false-positive rate $\le 0.10$ & 0.835 \\
\bottomrule
\end{tabular}
\end{table}

\subsection{Results on the real converted portion (JAAD-DVS)}\label{sec:results-jaad}

The benchmark's real portion consists of JAAD dash-cam video
converted to events
and is substantially harder than the synthetic portion: published
supervised results on these data range from near-chance to 0.70
AUROC depending on model and condition \cite{sakhai2024}. We
evaluate three configurations of the unsupervised pipeline on the
real test split, in increasing order of access to real data:
(A)~\emph{zero-shot transfer}, in which both the dictionary and the
readout are taken from the synthetic domain without modification;
(B)~\emph{readout adaptation}, in which the synthetic-trained
dictionary is kept frozen and only the linear readout is refitted on
the real training labels; and (C)~\emph{real-data training}, in
which the dictionary itself is learned, without labels, from the
real training frames and the readout is fitted on the real training
labels. Table~\ref{tab:jaad} reports AUROC and balanced accuracy for
each configuration, partitioned by the weather attribute of each
scene.

\begin{table}[h]
\caption{JAAD-DVS test split ($n{=}7{,}409$; 38\% crossing),
partitioned by the per-video JAAD weather attribute (adverse $=$
rain$+$snow, $n{=}633$). AUROC / balanced accuracy.}\label{tab:jaad}
\begin{tabular}{@{}lcccc@{}}
\toprule
Configuration & Combined & Clear & Cloudy & Adverse \\
\midrule
(A) zero-shot & 0.534 / 0.503 & 0.546 / 0.506 & 0.510 / 0.501 & 0.558 / 0.489 \\
(B) readout adaptation & 0.669 / 0.601 & 0.660 / 0.605 & 0.674 / 0.594 & 0.685 / 0.609 \\
(C) real-data training & 0.662 / 0.596 & 0.664 / 0.603 & 0.649 / 0.584 & 0.698 / 0.603 \\
\bottomrule
\end{tabular}
\end{table}

Zero-shot transfer fails: the synthetic-trained pipeline scores near
chance in every environment (combined AUROC 0.534), which parallels
the behaviour of supervised models transferred to these data without
adaptation \cite{sakhai2024} and quantifies the domain gap
introduced by the conversion process and by differing viewpoint and
scene statistics. Refitting only the linear readout on the real
training labels recovers most of the attainable performance
(configuration B, combined AUROC 0.669), placing the label-free
features within the range reported for fully supervised models on
these data (0.55--0.70 AUROC \cite{sakhai2024}). Training the
dictionary itself on the real frames (configuration C) does not
improve on the transferred dictionary (combined 0.662); the two
configurations differ by less than 0.01 AUROC in every environment
except adverse weather, where configuration C attains the highest
value observed on the real portion (0.698). The practical
implication mirrors the synthetic-side finding of
Sec.~\ref{sec:results-tiers}: the patch-level features are largely
domain-general, and the domain-specific component of the task is
concentrated in the readout, which is the component that is cheapest
to adapt. Two caveats apply: the real-portion split is frame-level
and stratified, so frames from one video may appear in both the
training and test partitions, and all real-portion results are
single-seed.

\subsection{Computational cost}\label{sec:results-cost}

Dictionary training takes eight minutes on a single consumer GPU and
requires neither labels nor backpropagation. Encoding amounts to one
matrix product per frame, and the readout is a linear model that can
be trained in minutes and refitted on-device.

\section{Discussion}\label{sec:discussion}

\paragraph{Annotation requirements.} These results alter the cost
structure of simulation-based development of event perception.
Feature learning, the data-intensive stage, consumes only unlabelled
streams, which simulators and test vehicles produce as a by-product
of operation; labels are required only for the linear readout, whose
sample efficiency is considerably higher, and for evaluation. On
this benchmark the measured accuracy cost relative to end-to-end
supervision is below two percentage points.

\paragraph{Interpretation of the readout difference.} The readout
difference reported in Sec.~\ref{sec:results-tiers} suggests a
reinterpretation of earlier results: unsupervised spiking approaches
on event data have typically been evaluated through whole-sample WTA
codes and assignment readouts, and their reported weakness may
reflect that protocol rather than a limitation of local plasticity.
A patch dictionary with a pooled linear readout is subject to the
same biological constraints, in that the learning remains local,
online and label-free, but it is better matched to the structure of
visual data and recovers most of the gap to supervision.

\paragraph{Implications for deployment.} The pipeline partitions
naturally for embedded use. The dictionary is the component that
would reside on a neuromorphic device, and each element of the
learner is standard in spiking simulators
\cite{gewaltig2007,stimberg2019,davison2009,hazan2018} and
neuromorphic toolchains. Encoding requires one matrix product per
frame, and the readout is a linear model that can be refitted
on-device. The robustness under adverse weather (0.913 AUROC) and
the freely recalibratable operating point are further properties
relevant to application settings.

\paragraph{Limitations.} The benchmark's real portion is derived
from conventional JAAD video converted to events; native event recordings of
crossing scenes are not yet available in public benchmarks, so the
conversion process remains a potential confound for all results on
real data. Frame-level evaluation follows the benchmark protocol
but does not exploit event timing; clip-level and latency-aware
evaluation is left to future work. The pipeline results are reported
for a single seed (the seed variability of the same learner measured
on other benchmarks is $\sigma \approx 0.002$), and the supervised
reference shares the data but differs in architecture class. These
results establish the feasibility of annotation-free training; they
do not constitute a road-ready system.

\paragraph{Generality.} No component of the pipeline is specific to
pedestrians, and the annotation considerations discussed above apply
to other perception functions in which sensor data accumulates
faster than it can be labelled, including other road users, traffic
signage, and rail and industrial safety monitoring. The crossing
task combines high annotation ambiguity, since the label refers to
behavioural intent, with a high cost of failure, and is therefore a
setting in which removing the labelled-feature requirement is
particularly valuable.

\paragraph{Future work.} Four extensions follow directly from this
study. First, clip-level and latency-aware evaluation: the
benchmark's frame protocol does not exploit event timing, and a
detector that integrates evidence across a clip should trade latency
against the operating points of Table~\ref{tab:operating}
explicitly. Second, native event recordings: existing crossing
benchmarks, including DVS-PedX, derive their real portions from
converted conventional video, and a native recording campaign of
moderate scale would remove the conversion confound; given the
annotation-free training demonstrated here, it would require labels
only for an evaluation subset. Third, label-free domain adaptation:
Sec.~\ref{sec:results-jaad} adapts the readout with real labels,
and a natural next step is to adapt the dictionary itself using only
unlabelled real streams, which would extend the annotation-free
property of the method across the domain gap. Fourth, architectural
depth: the dictionary used here is a single layer, and a
convolutional STDP hierarchy \cite{kheradpisheh2018} could supply
mid-level features while keeping learning local and label-free. A
complementary extension is a parameter-free attention mechanism, in
the form of a channel-wise homeostatic gain that modulates neuron
excitability from running activity statistics; because such a
mechanism introduces no learned parameters, it preserves the
unsupervised character of the feature-learning stage.

\section{Conclusion}\label{sec:conclusion}

We have presented the first unsupervised results on a public
event-camera benchmark for pedestrian crossing detection. Features
learned entirely without labels support 90.3\% accuracy and 0.936
AUROC, within 0.007 AUROC of a fully supervised spiking network
evaluated on identical test data, and robustness under adverse
weather is largely retained (0.913 AUROC). Our analysis further
indicates that the accuracy gap previously reported for unsupervised
spiking perception on tasks of this kind is attributable primarily
to the readout protocol rather than to the local learning rule. On
the benchmark's real converted portion the same features transfer:
zero-shot evaluation is at chance, a readout refit on real labels
reaches 0.669 AUROC, within the published supervised range on these
data, and training the dictionary on the real frames provides no
additional benefit. We conclude that, on this benchmark, labelled
data is required chiefly for the readout stage, and in comparatively
small quantities.

\backmatter

\bmhead{Acknowledgements}
This work was carried out in partial fulfilment of the requirements
for the doctoral degree of H.T.\ at Alma Mater Europaea University.
The authors thank their colleagues at Alma Mater Europaea University
and AGH for valuable discussions and support, and the DVS-PedX team for the public
benchmark.

\section*{Statements and Declarations}

\bmhead{Funding}
The authors declare that no funds, grants, or other support were
received during the preparation of this manuscript.

\bmhead{Competing interests}
M.~Sakhai is a co-author of the DVS-PedX dataset publication
\cite{sakhai2026dvspedx}. The authors have no other relevant
financial or non-financial interests to disclose.

\bmhead{Ethics approval}
Not applicable. This study uses only the publicly available DVS-PedX
benchmark and involves no human participants or animals.

\bmhead{Data availability}
The DVS-PedX dataset analysed in this study is publicly available on
Zenodo (\url{https://doi.org/10.5281/zenodo.17030898}). Per-frame
test predictions supporting the reported results are available from
the corresponding authors on reasonable request.

\bmhead{Code availability}
All training and evaluation code is permanently archived on Zenodo
(\url{https://doi.org/10.5281/zenodo.22032266}, v1.0) and maintained
at \url{https://github.com/anomaitech/anomaitech-unsupervised_snn_pedx}.

\bmhead{Author contributions}
H.T.: conceptualisation, methodology, software, experiments, and
writing (original draft). M.S.: data curation (DVS-PedX benchmark)
and writing (review and editing). M.M.: supervision and writing
(review and editing). M.W.: supervision, methodology, and writing
(review and editing). All authors read and approved the final
manuscript.

\begin{appendices}

\section{Hyperparameters and reproducibility}\label{app:hyper}

Table~\ref{tab:hyper} lists the full configuration. Nothing was tuned
on the test split; probe regularisation is selected on a 10\% shuffled
hold-out of the training set ($C \in \{0.01, 0.1, 1\}$; selected
$C{=}0.01$). All results derive from logged runs with fixed seed 42;
per-frame test scores are retained and all figures are generated by
script from those logs. 

\begin{table}[h]
\caption{Configuration.}\label{tab:hyper}
\begin{tabular}{@{}lll@{}}
\toprule
Group & Parameter & Value \\
\midrule
Neuron & membrane leak $\tau_v$ & 20 steps \\
 & fixed threshold / increment / decay & 0.8 / 0.05 / 200 pres. \\
 & lateral inhibition & 2.0 \\
Encoding & timesteps $T$ / Poisson gain $\lambda$ & 40 / 0.5 \\
STDP & $\eta_+$ / $\eta_-$ / $w_{\max}$ & 0.01 / $10^{-4}$ / 1.0 \\
 & trace constants & 20 steps \\
 & row norm target & 15 \\
Pipeline & patch / stride / pooling & $9{\times}9$ / 3 / $4{\times}4$ sum \\
 & dictionary size $K$ / patches / epochs & 2048 / 300k / 3 \\
 & probe & logistic regression, $C{=}0.01$ \\
Data & frames train/test & 114{,}107 (60{,}000 used) / 24{,}454 \\
\bottomrule
\end{tabular}
\end{table}

\section{Evaluation protocol details}\label{app:protocol}

\paragraph{Synthetic split.} The 163{,}012 single frames derive from
the DVS-PedX CARLA sequences (normal + adverse weather) under the
frame protocol of the supervised reference: stratified 70/15/15 by label with
seed 42, giving 114{,}107 / 24{,}451 / 24{,}454 frames; the dictionary
and probe use only the training partition (60{,}000 frames for the
probe), the validation partition is unused by the pipeline (its C
selection uses a 10\% shuffled hold-out of the training set), and all
reported numbers are on the untouched test partition. Weather
partitions follow the sequence provenance (15{,}776 normal / 8{,}678
adverse test frames).

\paragraph{Metrics.} AUROC is computed by the Mann--Whitney statistic
with mid-rank ties on the probe's positive-class probability. Operating
points (Table~\ref{tab:operating}) sweep the same score. Per-class F1
uses the argmax decision. No metric involves the validation partition.

\paragraph{Compute.} Dictionary training: eight minutes (RTX 5060
Ti). Encoding: $\sim$3 minutes for all 187k synthetic frames.
Probe fits: 15--40 minutes each (CPU). End-to-end, a full
from-scratch reproduction of every number in this paper is under
four hours on one consumer workstation.

\end{appendices}

\bibliography{sn-bibliography}

\end{document}